\documentclass[sigconf]{acmart} 
\copyrightyear{2026}
\acmYear{2026}
\setcopyright{cc}
\setcctype{by}
\acmConference[ICSE '26]{2026 IEEE/ACM 48th International Conference on Software Engineering}{April 12--18, 2026}{Rio de Janeiro, Brazil}
\acmBooktitle{2026 IEEE/ACM 48th International Conference on Software Engineering (ICSE '26), April 12--18, 2026, Rio de Janeiro, Brazil}
\acmPrice{}
\acmDOI{10.1145/3744916.3773115}
\acmISBN{979-8-4007-2025-3/26/04}

\author{Vishnu Asutosh Dasu}
\affiliation{%
  \institution{Pennsylvania State University}
  \city{State College}
  \country{USA}
}
\email{vdasu@psu.edu}

\author{Md Rafi ur Rashid}
\affiliation{%
  \institution{Pennsylvania State University}
  \city{State College}
  \country{USA}
}
\email{rafiurrashid@psu.edu}

\author{Vipul Gupta}
\affiliation{%
  \institution{Pennsylvania State University}
  \city{State College}
  \country{USA}
}
\email{vkg5164@psu.edu}

\author{Saeid Tizpaz-Niari}
\affiliation{%
  \institution{University of Illinois}
  \city{Chicago}
  \country{USA}
}
\email{saeid@uic.edu}

\author{Gang Tan}
\affiliation{%
  \institution{Pennsylvania State University}
  \city{State College}
  \country{USA}
}
\email{gtan@psu.edu}

\usepackage{amsthm}
\usepackage{amsmath}
\usepackage{centernot}
\usepackage{comment}
\usepackage{microtype}
\usepackage{siunitx}
\usepackage{multirow}
\usepackage{subcaption}

\DeclareMathOperator*{\argmin}{arg\,min}

\newcolumntype{Y}{>{\centering\arraybackslash\hsize=1\hsize}X}
\newcolumntype{C}{>{\centering\arraybackslash\hsize=3\hsize}X}

\newcolumntype{P}{>{\centering\arraybackslash}p{0.1\textwidth}}
\newcolumntype{R}{>{\centering\arraybackslash}p{0.1\textwidth}}
\newcolumntype{T}{>{\centering\arraybackslash}X}
\newcolumntype{S}{>{\centering\arraybackslash}p{0.1\textwidth}}

\newcommand{\st}{\scriptscriptstyle}

\newcommand{\revision}[1]{\textcolor{black}{#1}}

\newcommand\vd[2]{d_{i, p}}

\newcommand{\toolname}{\textsc{AP}\xspace}
\newcommand{\toolnamefull}
{\textsc{Attention Pruning}\xspace}

\usepackage{tikz}

\newtheorem{definition}{Definition}[section]

\definecolor{gold}{rgb}{0.99,0.78,0.07}
\usetikzlibrary{arrows,shapes,snakes,automata,backgrounds,positioning,decorations.pathmorphing,
	decorations.markings,calc}

\tikzstyle{dtreenode}=[draw=blue!10!gray,rounded rectangle, minimum size=5mm,fill=blue!10!white]
\tikzstyle{dtreeleaf}=[draw=black!60,minimum width=1cm,minimum height=0.4cm,rectangle,fill=blue!50!white]
\tikzset{every loop/.style={looseness=7}}
\tikzset{
	gluon/.style={decorate,draw=black,
		decoration={coil,amplitude=1pt, segment length=5pt}}
}
\tikzset{
	gluon1/.style={decorate,draw=black,
		decoration={coil,amplitude=3pt, segment length=3pt}}
}
\tikzset{
	gluonew/.style={decorate,draw=black,
		decoration={coil,amplitude=1pt, segment length=2pt}}
}

\tikzset{bicolor/.style args={#1 and #2 and #3}{
		path picture={
			\tikzset{rounded corners=0}
			\fill [#1] (path picture bounding box.south west)
			rectangle
			($(path picture  bounding box.north west)!#3!(path picture bounding
			box.north east)$);
			\fill [#2]
			($(path picture bounding box.south west)!#3!(path picture bounding
			box.south east)$)
			rectangle (path picture bounding box.north east);
}}}

\tikzset{tricolor/.style args={#1 and #2 and #3 and #4 and #5}{
		path picture={
			\tikzset{rounded corners=0}
			\fill [#1] (path picture bounding box.south west)
			rectangle
			($(path picture  bounding box.north west)!#4!(path picture bounding
			box.north east)$);
			\fill [#2]
			($(path picture bounding box.south west)!#4!(path picture bounding
			box.south east)$)
			rectangle
			($(path picture  bounding box.north west)!#5!(path picture bounding
			box.north east)$);
			\fill [#3]
			($(path picture bounding box.south west)!#5!(path picture bounding
			box.south east)$)
			rectangle (path picture bounding box.north east);
}}}

\usepackage[many]{tcolorbox}
\tcbuselibrary{listings,skins}

\lstdefinestyle{mystyle}{
  xleftmargin=0pt,
   basicstyle={\footnotesize\ttfamily},
   aboveskip=3mm,
   belowskip=3mm,
   keywordstyle=\bfseries,
   showstringspaces=false,
  escapechar=?,
  language=Java
}
\definecolor{code_indent}{HTML}{CCCCCC}

\newtcblisting{mylisting}[2][]{
  arc=0pt, outer arc=0pt,
  listing only,
  listing style=mystyle,
  title={\large #2},
  #1
}

 \definecolor{dkgreen}{rgb}{0,0.6,0}
 \definecolor{gray}{rgb}{0.5,0.5,0.5}
 \definecolor{mauve}{rgb}{0.58,0,0.82}

\definecolor{cadmiumgreen}{rgb}{0.0, 0.42, 0.24}
\definecolor{verde}{rgb}{0.25,0.5,0.35}
\definecolor{jpurple}{rgb}{0.5,0,0.35}
\definecolor{darkgreen}{rgb}{0.0, 0.2, 0.13}

\usepackage[ruled,vlined,linesnumbered,lined]{algorithm2e}
\usepackage{algpseudocode}

\usepackage{mathtools,xparse}

\usepackage{tabu}
\usepackage{multirow}

\usepackage{pifont}
\usepackage{enumerate}
\usepackage[shortlabels]{enumitem}

\usepackage{pgfplotstable}
\usepackage{pgfplots}

\usepackage[noframe]{showframe}
\usepackage{framed}

 {\endMakeFramed}
 \definecolor{shadecolor}{gray}{0.85}

\definecolor{bgblue}{RGB}{245,243,253}
\definecolor{ttblue}{RGB}{91,194,224}

\newtcolorbox{myboxi}[1][]{
  breakable,
  title=#1,
  colback=white,
  colbacktitle=white,
  coltitle=black,
  fonttitle=\bfseries,
  bottomrule=0pt,
  toprule=0pt,
  leftrule=3pt,
  rightrule=3pt,
  titlerule=0pt,
  arc=0pt,
  outer arc=0pt,
  colframe=black!50,
}

\newtcolorbox{myboxii}[1][style=mystyle]{
  breakable,
  freelance,
  colback=white,
  colbacktitle=white,
  coltitle=black,
  fonttitle=\bfseries,
  bottomrule=0pt,
  boxrule=0pt,
  colframe=white,
  after skip=0pt,
  overlay unbroken and first={
    \draw[white!75!black,line width=3pt]
    ([yshift=-9pt]frame.north west) --
    ([yshift=9pt]frame.south west);
  },
  }

\begin{document}

\title{\textsc{Attention Pruning}: Automated Fairness Repair of Language Models via Surrogate Simulated Annealing}

\renewcommand{\shortauthors}{Dasu et al.}

\begin{abstract}
This paper explores pruning attention heads as a post-processing bias mitigation method for large language models (LLMs).  
LLMs have been expanding into sensitive social contexts and socio-economic decision-making where fairness concerns become especially crucial.
Since LLMs develop their decision-making patterns by training on massive datasets of human-generated content, they naturally encode and perpetuate societal biases.
While modifying training datasets and algorithms is prohibitive, post-processing techniques---such as pruning attention heads in pre-trained LLMs---can provide feasible and effective approaches to improve fairness. However, identifying the optimal subset of parameters to prune presents a combinatorial challenge within the immense parameter space of LLMs, requiring efficient solutions that balance competing objectives across the frontiers of model fairness and utility. 

We explore a search-based program repair approach via simulated annealing to address the computational challenges. Given the prohibitive evaluation costs in billion-parameter LLMs, we develop surrogate deep neural networks that efficiently model the relationship between attention head states (active/inactive) and their corresponding fairness/utility metrics. This allows us to perform optimization over the surrogate models and efficiently identify optimal subsets of attention heads for pruning rather than directly searching through the LLM parameter space. This paper introduces \toolnamefull, a fairness-aware surrogate simulated annealing approach to prune attention heads in LLMs that disproportionately contribute to bias while minimally impacting overall model utility. Our experimental evaluation shows that \toolnamefull achieves a reduction of up to $40\%$ in gender bias and outperforms state-of-the-art bias mitigation strategies.  

\textcolor{red}{Warning: This paper contains content that some readers may find offensive and harmful.}


\end{abstract}




\setcopyright{none} 
\settopmatter{printacmref=false} 

\maketitle

\section{Introduction}

\revision{Recent advances in large language models (LLMs) have enabled near-human performance on various tasks \cite{openai2024gpt4technicalreport} and have catalyzed their widespread adoption in real-world use cases.} However, despite their impressive performance, prior work has shown that these models often inherit biases from training data \cite{hada-etal-2023-fifty}. 
To make matters worse, such biases can produce uneven performance across different social groups in society \cite{gupta2023calm, li-etal-2020-unqovering,nagireddy2024socialstigmaqa, guo2024hey}.
Bias in language models is particularly troubling given its potential for significant negative societal impacts \cite{UNESCO2024, Milne2024,gallegos2024bias}.
Consequently, equipping LLM-based software developers to mitigate bias in socially critical AI systems is essential to promote inclusivity and equality.

Significant research within the software engineering (SE) community has focused on addressing discrimination in data-driven software systems~\cite{FairnessTesting, 10.1145/3338906.3338937,udeshi2018automated,ADF,chakraborty2020fairway}. Fairness has emerged as a crucial meta-functional property, requiring analysis beyond functional correctness and evaluation metrics that go beyond prediction accuracy~\cite{brun2018software}. Consequently, researchers have developed various testing~\cite{FairnessTesting, 10.1145/3338906.3338937,ADF}, debugging~\cite{10.1145/3468264.3468536, 10.1109/ICSE48619.2023.00136, yu2024fairlay}, and repairing~\cite{10.1145/3650212.3680380,10.1109/TSE.2022.3220713,chen2022maat,10.1145/3617168,10.1145/3510003.3510087} approaches to address fairness issues in data-driven software systems. However, repairing unfairness in the frontier AI models has been understudied within the SE community.  

Various techniques have been proposed to mitigate social biases in language models within the AI community. Some past works aim to find a subset of weights containing bias and then modify only these weights to improve fairness~\cite{meissner-etal-2022-debiasing, le2020adversarial}. Others tried to unlearn the undesired properties by modifying the representations \cite{li2024wmdp}. 
Recently, Zayed et al.~\cite{zayed2024fairness} presented a pruning strategy to mitigate unfairness in LLMs that outperformed prior works. Despite all these efforts, improving fairness in LLMs while balancing the inevitable trade-off in model utility has remained a challenge. 

Inspired by search-based program repair techniques ~\cite{10.1145/2568225.2568254,10.1145/1569901.1570031,ruan2024evolutionary}, we propose a randomized search algorithm to semantically repair unfairness in LLMs without degrading their functional utilities. 
Our key observation is that attention heads in LLMs may disparately contribute to unfairness while minimally affecting utility. Hence, pruning an ideal set of attention heads as a post-processing mitigation strategy can improve fairness with a negligible impact on utility. However, pruning is a combinatorial search problem that requires an exhaustive search over all possible subsets of attention heads and poses a significant computation challenge in LLMs with over a thousand attention heads. 
Prior work on fairness-aware attention head pruning \cite{zayed2024fairness} avoids the combinatorial problem by only analyzing the effect each attention head has on fairness. Such an approach fails to account for different combinations of attention heads and their complex nonlinear relationships.

We propose a randomized heuristic search in the class of simulated annealing (SA) algorithms to explore the search space efficiently. 
However, due to the high inference time of billion-parameter LLMs, a direct application of SA is infeasible, as exploring even a single state takes several minutes.
We make a key observation that the effect of pruning a subset of attention heads on fairness/utility follows a pattern that can be learned using \emph{surrogate functions} via deep neural networks (DNNs). This novel use of surrogate functions scales the SA search to find an ideal subset of attention heads using the surrogate models rather than the LLMs. 

We implement and evaluate \toolnamefull (\toolname), an automated fairness repair strategy for LLMs using SA. \toolname identifies an ideal subset of attention heads to prune such that the fairness of LLMs improves while ensuring minimal loss in model utility. It uses surrogate DNNs during the SA search to explore the combinatorial search space of billion-parameter LLMs efficiently.  We investigate the following research questions with 6 LLMs, focusing on reducing gender bias from the HolisticBias~\cite{smith2022imsorryhearthat} dataset:

\vspace{0.25 em}
\noindent \emph{RQ1) Effectiveness of surrogate DNNs.} We observe that surrogate DNNs accurately predict the effect of pruning attention heads on the fairness and utility with low mean squared error ($\text{MSE} \le 0.01$). 

\vspace{0.25 em}
\noindent \emph{RQ2) Comparison to the state-of-the-art methods.}
Our experiments show that \toolname effectively uses surrogate SA and outperforms the state-of-the-art fairness-aware pruning strategies. We observe that unfairness reduces by up to $40\%$ and in $4/6$ cases \toolname finds states that have better fairness and utility than the state-of-the-art. 

\vspace{0.25 em}
\noindent \emph{RQ3) Design considerations of \toolnamefull.} We examine how different hyperparameters can be tuned to have fine-grained control over the fairness-utility trade-off. We identify one hyperparameter that can effectively explore the Pareto frontier of fairness-utility in LLMs and balance their trade-off. 

\vspace{0.25em}
\noindent \emph{RQ4) Generalization beyond gender bias.} We show that reducing gender bias with \toolname also reduces other forms of social biases in LLMs. We experiment with race, nationality, sexual orientation, and age bias and observe up to 65\% improvement in fairness. Additionally, we also show that \toolname can also directly reduce other biases like race.

In summary, the key contributions of this paper are:
\begin{enumerate}
    \item \textbf{Approximating the Fairness and Utility of LLMs.} We show that the fairness and utility of pruned LLMs can be accurately approximated using small surrogate deep neural networks.
    \item \textbf{Surrogate Simulated Annealing.} We propose an efficient discrete optimization algorithm using surrogate simulated annealing that handles billion-parameter LLMs and outperforms the state-of-the-art bias mitigators. 
    \item \textbf{Experimental Evaluation.} We present \toolname and evaluate its effectiveness in reducing social biases in various LLMs.    
\end{enumerate}

\section{Background}
\subsection{Causal Language Modeling}

Causal Language Modeling (CLM) is a natural language processing task where the LLM aims to predict the next word or token given a sequence of tokens. The LLM autoregressively generates the next token until a pre-determined sequence length is reached or a special STOP token is generated. Given a sequence of $n$ tokens $\mathbf{y} = \{y_{\st 1}, y_{\st 2} \ldots, y_{\st n-1}, y_{\st n}\}$, the language model $\Theta$ is trained to learn the following probability distribution:
\begin{equation}
    \label{eq:clm_prob}
    P(\mathbf{y}) = {\displaystyle \prod_{\st i=1}^{\st n} P(y_{\st i}|y_{\st 1}, \ldots, y_{\st i-1})}
\end{equation}

The CLM training objective is to get a model $\Theta$ to minimize the negative log-likelihood loss given by:
\begin{equation}
\mathcal{L}(\Theta,\mathbf{y}) = -\displaystyle \sum_{\st i=1}^{\st n} \log(\Theta(y_{\st i}|y_{\st 1}, \ldots, y_{\st i-1}))
\end{equation}

After training is complete, the text is autoregressively sampled from the language model; i.e., $\hat{y}_{\st t < n} \sim \Theta(y_{\st t} | y_{\st 1}, \ldots, y_{\st t-1})$.

\subsection{Bias in Causal Language Modeling}

Bias in LLMs for causal language modeling can be broadly categorized into two categories \cite{guo2024biaslargelanguagemodels}: \textit{intrinsic bias} and \textit{extrinsic bias}. Intrinsic biases arise from the training dataset, learning algorithms, or the model architecture itself \cite{sun-etal-2019-mitigating}. LLMs trained on vast text sequences from the Internet tend to inherit the biases present in them. For example, LLMs reinforce gender stereotypes by associating the word ``doctor'' with men and ``nurse'' with women \cite{bolukbasi2016mancomputerprogrammerwoman}.
The side effects of such biases are amplified when they are employed in sensitive socio-economic applications.





Extrinsic biases in LLMs manifest in specific real-world applications. As a result, the metrics and benchmarks that have been proposed to evaluate extrinsic bias are task-specific. For example, \citet{sap-etal-2019-risk} shows that tweets written in African American English (AAE) by self-identified African-Americans are two times more likely to be labeled as offensive than other tweets. 
Other benchmarks \cite{rudinger2018genderbiascoreferenceresolution} contains a set of Winograd-schema style sentences for coreference resolution focused on gender bias. For example, given the sentence ``the doctor asked
the nurse to help her because she was busy'', models often erroneously correlate ``her'' with ``nurse'' rather than with ``doctor''. 


\subsection{Notions of Model Utility via Perplexity}

The utility of a language model is commonly measured through the \textit{perplexity} metric, which measures how well the model has learned the probability distribution in Equation~\ref{eq:clm_prob}. The perplexity is computed from the causal language modeling loss function with respect to a fixed sequence of tokens $\mathbf{y}$. Formally, the perplexity $PPL$ of a sequence $\mathbf{y}$ is defined as
\begin{equation}
\label{ppl}
    PPL_{\Theta}(\mathbf{y}) = \exp\left(-\frac{1}{n}{\displaystyle \sum_{\st i=1}^{\st n} \log(\Theta(y_{\st i}|y_{\st 1}, \ldots, y_{\st i-1}))}\right)
\end{equation}

A lower perplexity score implies that model $\Theta$ has been trained well to estimate the real-world probability distribution. Informally, a sequence with a low perplexity score implies that the model is less ``surprised'' by a sequence of tokens. In our work, we measure the perplexity on the WikiText-2 \cite{merity2016pointersentinelmixturemodels} dataset.

\subsection{Notions of Model Fairness}
In our work, we use the HolisticBias \cite{smith2022imsorryhearthat} metric that measures a form of extrinsic bias. The HolisticBias metric builds on BOLD \cite{Dhamala_2021} and contains a dataset of 566,000 prompts that are categorized into 13 social biases (e.g., Gender, Sexual Orientation, Race and Ethnicity, Nationality, etc.). Bias is quantified by prompting the LLMs with sentences from the different groups and examining the toxicity of the generated text.  The toxicity of a text determines how disrespectful, unpleasant, or harmful the language is. The toxicity is computed using a BERT \cite{devlin2019bert} model that is pre-trained on a toxic comment classification dataset\footnote{\url{https://www.kaggle.com/c/jigsaw-toxic-comment-classification-challenge}}. The BERT model outputs a score in $[0,1]$, where a higher score is assigned to more toxic text. 

To measure the bias of the LLM, we adopt the procedure outlined in prior works \cite{zayed2024fairness,Dhamala_2021}. Each social bias, or group, in the HolisticBias dataset consists of a set of subgroups $G$. For example, the gender group contains subgroups such as binary, queer, transgender, etc. 
The bias of the LLM $\Theta$ on a set of subgroups $G$ is

\begin{equation}
\text{bias}_\Theta(G) = \sum_{g \in G} \left| T_G - T_g \right|
\end{equation}

\begin{equation}
\text{where} \quad T_g = \frac{1}{|D_g|} \sum_{x \in D_g} \text{tox}_\Theta(x)
\end{equation}

\begin{equation}
\text{and} \quad T_G = \frac{1}{|G|} \sum_{g \in G} T_g
\end{equation}

In the above equations, $T_g$ is the toxicity of a subgroup $g \in G$, and $T_G$ is the average toxicity of all subgroups. $D$ is the set of HolisticBias prompts and $D_g \in D$ are the prompts that belong to subgroup $g$.  The toxicity of the text generated by LLM $\Theta$ when prompted with $x$ is denoted by $tox_{\Theta}(x)$. Lower bias values indicate less bias in the LLM. We note that in HolisticBias, the bias is related to toxicity but is still fundamentally different. Low average toxicity does not imply low bias, as bias is computed as the sum of the differences in subgroup toxicity and overall average group toxicity.

\section{Problem Statement}
Our work focuses on repairing unfairness in LLMs by removing unfair attention heads. This section describes the attention mechanism and formally defines the problem statement.

\subsection{Multi-Head Attention in Transformers}
Most modern LLMs for causal language modeling are based on the Transformer architecture proposed by \citet{10.5555/3295222.3295349}. On a high level, the Transformer is a neural network that comprises $n$ blocks stacked on top of each other, where each block has a Multi-Head Attention (MHA) layer followed by a fully connected layer. The MHA layer utilizes scaled-dot product attention, which operates on a set of queries $Q$, keys $K$, and values $V$, which come from the previous layer's output. The query, key, and values are computed from the input text embeddings for the first block. Scaled-dot product attention is computed as
\begin{equation}
    \text{Attention}(Q,K,V) = \text{softmax}\left( \frac{QK^T}{\sqrt{d_k}}
    \right)V
\end{equation}

where $d_k$ is the dimension of the keys. MHA comprises $h$ scaled-dot product attention heads concatenated and multiplied with an output projection matrix. Formally, this is represented as
\begin{equation}
    MultiHead(Q,K,V) = Concat(head_1,\ldots,head_n)W^O
\end{equation}

where $head_i$ is computed as 
\begin{equation}
    head_i = \text{Attention}(QW_i^Q, KW_i^K, VW_i^V)
\end{equation}

and $W_i$ and $W^O$ are learned parameter matrices. Figure~\ref{fig:mha_original} shows an MHA layer in a Transformer block with four heads. 


\subsection{Attention Head Pruning for Fairness}
\begin{figure}[!tbh]
    \centering
    \begin{subfigure}{0.22\textwidth}
        \centering
        \includegraphics[width=\linewidth]{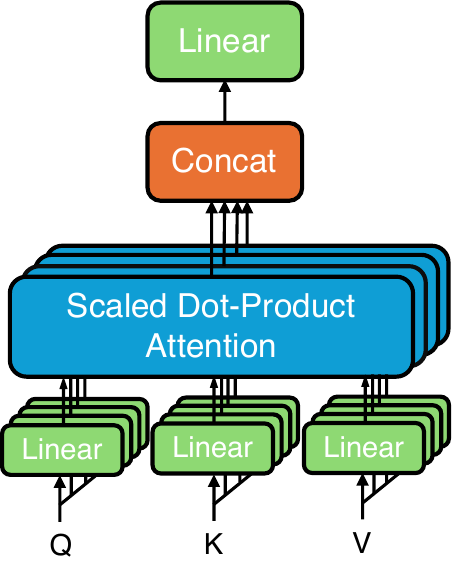}
        \caption{Multi-Head Attention with four attention heads before pruning}
        \label{fig:mha_original}
    \end{subfigure}
    \hfill
    \begin{subfigure}{0.22\textwidth}
        \centering
        \includegraphics[width=\linewidth]{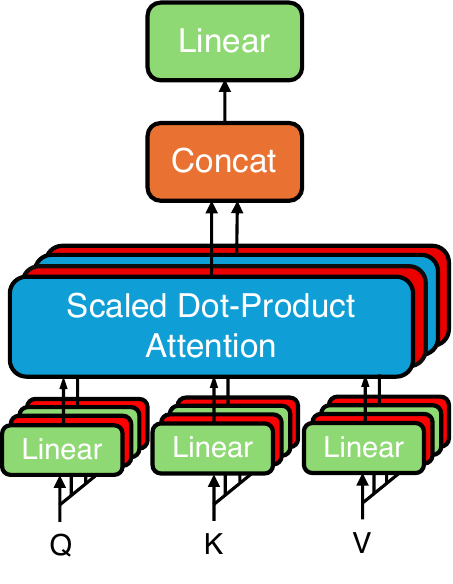}
        \caption{Multi-Head Attention after pruning the second and fourth heads}
        \label{fig:mha_pruned}
    \end{subfigure}
    \caption{Multi-Head Attention in a Transformer block with four attention heads before and after pruning}
    \label{fig:main}
\end{figure}

Pruning in LLMs refers to the practice of ``pruning'' or ``dropping'' parts of the LLM. Pruning strategies can be broadly categorized into two categories: \textit{unstructured} and \textit{structured} pruning. Unstructured pruning refers to the practice of pruning at the neuron level, where a subset of neurons are pruned. On the other hand, structured pruning strategies remove entire structural components, such as attention heads or layers, from the LLM. Attention head pruning is a structured pruning strategy that prunes one or more attention heads from multi-head attention in a Transformer block. For example, Figure~\ref{fig:mha_pruned} shows multi-head attention with the second and fourth heads pruned. These heads are not concatenated before multiplying with the output projection matrix. 

Multiple works have been proposed \cite{DBLP:journals/corr/abs-1905-10650,voita-etal-2019-analyzing,fan2021layerwisemodelpruningbased,han2016deepcompressioncompressingdeep,behnke-heafield-2020-losing} to prune redundant attention heads in LLMs that have a negligible impact on utility. However, they do not account for fairness. We define the problem of pruning attention heads to improve fairness while minimizing loss to model utility as follows


\begin{definition}[Fairness-Aware Attention Head Pruning]\label{def:problem}
Given an LLM model $\Theta$ with a set of attention heads $H$, a utility notion $PPL({\Theta})$, and a bias notion $Bias({\Theta})$, the fairness repair problem is to find a subset of attention heads $H' \subset H$ for pruning such that the repaired LLM model $\Theta' = \Theta \setminus H'$ minimizes the bias with the least degradation of model utility. Formally,
\[
\text{Find } \Theta' = H \setminus H' \text{ that min. } \epsilon \cdot Bias(\Theta') + (1 - \epsilon) \cdot (PPL(\Theta') - PPL(\Theta)),
\]
where $\epsilon \in [0, 1]$ is a weighting factor that trades-off between minimizing bias $Bias(\Theta')$ and minimizing utility degradation $PPL(\Theta') - PPL(\Theta)$, where $PPL(\Theta') - PPL(\Theta) \geq 0$ represents the utility loss. 
\end{definition}

Recently, \citet{zayed2024fairness} proposed the fairness-aware structured pruning (FASP) algorithm that prunes attention heads while accounting for both fairness and model utility. FASP quantifies the contribution of an attention head by measuring the change in bias and perplexity of LLM before and after pruning it. Formally, for a model with $N$ attention heads, the impact of each head $h$ on a social group represented by set $G$ is estimated as
\begin{equation}
z_{bias}(h,G) = bias_{\Theta}(G) - bias_{\Theta \setminus h}(G)
\end{equation}


where $bias_{\Theta}(G)$ and $bias_{\Theta \setminus h}(G)$ are the biases of the text generated by LLM $\Theta$ before and after pruning head $h$.  

Similarly, the effect of head $h$ on perplexity is computed as
\begin{equation}
    z_{ppl}(h) = ppl_{\Theta} - ppl_{\Theta \setminus h}
\end{equation}
where $ppl_{\Theta}$ and $ppl_{\Theta \setminus h}$ are the perplexities the LLM $\Theta$ before and after pruning head $h$.

Let the set of all heads be $H$. The FASP algorithm first sorts the attention heads in non-increasing order based on $z_{ppl}$. Then, it identifies the set of critical heads $H_C$ by isolating the top $\gamma \times N$ heads. Here, $\gamma \in [0,1]$ is a hyperparameter that determines the ratio of heads that are critical for model utility. The remaining $H \setminus H_C$ heads are then sorted in non-increasing order based on $z_{bias}$. Finally, the top $\alpha \times N$ heads are pruned from $H \setminus H_C$ based on their effect on $z_{bias}$. Here, $\alpha \in [0,1]$ is a hyperparameter that determines the ratio of attention heads that must be pruned. These pruned heads are deemed important by FASP for bias reduction but not critical for perplexity.

As the authors concede, FASP fails to account for the complex non-linear relationship of attention heads in an LLM. 
For example, if dropping the first and second heads decreases bias while not reducing perplexity, dropping both of them together may not decrease bias, and it might worsen perplexity.
Nevertheless, FASP is a first step toward fairness-aware attention head pruning.

\section{Approach}
We propose \toolnamefull (\toolname), a fairness-aware attention head pruning algorithm that accounts for the complex non-linear relationships among attention heads. 

\subsection{Discrete Optimization with Simulated Annealing}

Given an LLM $\Theta$ with a set of attention heads $H = \{1,\ldots,N\}$, we aim to find a subset of heads $H'$ that minimize bias while retaining model utility. We pose this as a discrete optimization problem over a search space of size $2^N$, where $N = |H|$. 
We use the Simulated Annealing (SA) randomized algorithm to efficiently explore the exponentially large search space to find an approximate solution. 
We can formulate our discrete optimization problem using SA if we define the following:

\paragraph{Search Space} The search space $\mathcal{S}$ can be represented as an $N-$dimensional Boolean hypercube. Every subset of heads $s \subseteq H$ can be represented as a bitstring of length $N$, where the $i^{th}$ bit is set to 1 if the corresponding attention head is in $s$. The bitstrings can be interpreted as bitmasks for deciding which heads to prune.

Formally, the search space $\mathcal{S}$ is defined as a graph
\begin{equation}
    \mathcal{S} = (V, E)
\end{equation}
where the set of vertices \( V \) consists of all bitstrings of length \( N \)
\begin{equation}
\label{vertices}
    V = \{0,1\}^N
\end{equation}
and the set of edges \( E \) connects pairs of vertices that differ in exactly one bit (i.e., have Hamming distance 1)
\begin{equation}
\label{edges}
      E = \{(x, y) \in V \times V : d_H(x, y) = 1\}
\end{equation}

Here, \( d_H(x, y) \) is the Hamming distance between the bitstrings \( x \) and \( y \), defined as
\begin{equation}
  d_H(x, y) = \sum_{i=1}^{N} \mathbf{1}_{x_i \neq y_i}
\end{equation}

Two adjacent vertices, or \textit{states}, in $\mathcal{S}$, denote two subsets of heads $s, s'$ that differ in only one head. In other words, $s,s' \in H$ such that $|s \setminus s'| = 1$ or $|s' \setminus s| = 1$. 

\paragraph{Neighborhood Relation}
The neighborhood relation in SA defines how the search space is explored. The neighborhood of a state (or vertex) defines the set of states that are connected to it. With our definition of edges in our search space in Equation~\ref{edges}, the neighborhood of a state is the set of all states that are at a Hamming distance of one from it. Formally, this is denoted as
\begin{equation}
     \Gamma(s) = \{s' \in \mathcal{S} \; : \; d_H(s,s') = 1 \}
\end{equation}
To generate a new state from the current state $s$ and explore the search space, we randomly sample a state $s' \in \Gamma(s)$. The initial state $s_0$ is randomly sampled from the state space $\mathcal{S}$. In our SA search, we introduce two hyperparameters---$n_l$ and $n_u$---that control the number of pruned attention heads. The hyperparameter $n_l$ sets the lower bound on the number of heads that are pruned, and $n_u$ sets the upper bound. Setting bounds for the number of heads that are pruned is useful, as pruning too many or too few heads does not achieve the desired effect of low bias and low perplexity. \revision{Furthermore, while SA will converge without setting bounds on the number of heads pruned, it will take much longer as the search space is bigger.} The restricted bounded neighborhood is defined as 
\begin{equation}
     \Gamma'(s) = \{s' \in \Gamma(s) \; : \; n_l \le HW(s') \le n_u \}
\end{equation}
where $HW(\cdot)$ returns the Hamming weight of a state. 

\paragraph{Cost Function}
The cost function in SA assigns a score to each state $s \in S$ that defines how ``good'' the state is. Our goal is to find a state that represents the set of attention heads to be pruned that minimizes bias while retaining the perplexity of the model. We need states that have both low bias and low perplexity. Therefore, we propose the following linear weighted cost function that combines bias and perplexity to measure the quality of a state 
\begin{equation}
    \text{cost}(s) = \epsilon\cdot bias_{\Theta\setminus s} + (1-\epsilon)\cdot ppl_{\Theta \setminus s}
\end{equation}

where $\epsilon \in [0,1]$, $bias_{\Theta \setminus s}$ and $ppl_{\Theta \setminus s}$ are the bias and perplexity of the LLM $\Theta$ after pruning the attention heads in $s$. The hyperparameter $\epsilon$ balances the bias/perplexity trade-off during the search. An $\epsilon = 0.5$ assigns an equal weight to both bias and perplexity.

\paragraph{Temperature}

The temperature $T$ is a hyperparameter that affects the transition probability during state exploration of Simulated Annealing (SA). The transition probability to a state that has a higher cost than the current state is given by
\begin{equation}
    \text{Transition Probability} = e^{-\Delta E/T_i}
\end{equation}

where $\Delta E$ is the cost difference between the current and new state and $T_i$ is the temperature at the $i^{th}$ iteration. Loosely speaking, the temperature controls the exploration and exploitation trade-off in SA. For a fixed cost difference, a higher temperature increases the transition probability, thereby encouraging search space exploration. On the other hand, a lower temperature decreases the transition probability and encourages SA to restrict the search to states closer to states with a low cost.

\paragraph{Cooling Schedule} The cooling schedule determines how the temperature changes over time. In SA, we start with a high temperature and \textit{anneal} it over time. This encourages exploration in the early stages of the SA search and exploitation towards the end. We adopt the logarithmic cooling schedule \cite{hajek1988cooling}. The temperature $T_i$ at the $i^{th}$ iteration is defined as
\begin{equation}
\label{eq:cooling_schedule}
    T_i = \frac{T_0}{\log(2 + i)}, \forall i \in \mathbb{Z}_{\ge 0}
\end{equation}

where $T_0$ is the initial temperature. We utilize the temperature initialization algorithm proposed by \citet{Ben-Ameur2004} to set $T_0$. 

Although randomized algorithms like SA are faster than exhaustive search over combinatorial spaces, the high inference time of LLMs still makes them infeasible.
For example, one round of inference to compute the bias and perplexity of the Llama-2-7B model with 1024 attention heads takes $\approx13$ minutes on the NVIDIA RTX A6000 GPU. SA is inefficient in exploring a search space with $2^{1024}$ states as in one hour; we can only explore $\approx5$ states. To effectively apply SA for pruning attention heads, we need to scale up the search to explore thousands of states per second. Our solution is to train and leverage small DNNs as surrogate functions to approximate the fairness-utility frontiers of LLMs. 



\subsection{Approximating Bias and Perplexity of Pruned Large Language Models}

We need to speed up the search space exploration process to effectively utilize SA for discrete optimization in a reasonable amount of time. To achieve this speed-up, we leverage a key insight that we discover: \emph{the effect of pruning a subset of attention heads on the bias and perplexity of LLMs is not completely random and can be predicted using regression algorithms}. In other words, there is a pattern in attention head pruning that machine learning algorithms can learn. Our solution, therefore, is to use two \textit{surrogate} DNNs to predict the bias and perplexity of a state instead of directly using the LLM.

Algorithm~\ref{alg:dnn_train} outlines the training process for the surrogate DNNs. The first step of the training process is dataset creation.  We sample random subsets of the HolisticBias validation prompts and text from the validation data of WikiText-2. Random subsets are sampled instead of using the entire dataset for efficiency purposes. 
We pick a random state $s \in \mathcal{S}$ and prune the LLM $\Theta$. The pruned LLM $\Theta'$ is then prompted with the subset of HolisticBias prompts, and the bias is calculated. Similarly, the perplexity of $\Theta'$ is calculated on the subset of WikiText-2. The sampling and pruning process to create the dataset is repeated until a predefined time limit is reached. The second step is training the surrogate DNNs on the created dataset. We formulate the training objective as a regression task to predict the bias/perplexity by minimizing the mean squared error. Table~\ref{t:dataset_samples} presents a few samples from the training dataset used to train the surrogate DNNs for the Llama-2-7B and GPT-J-6B LLMs. The \textit{Attention Head Configuration} column represents a state in the search space. 
A ``1'' in the $i^{th}$ position of the bit string means that the $i^{th}$ attention head is pruned. 
The \textit{Bias} and \textit{Perplexity} represent the actual bias and perplexity observed by pruning the corresponding attention heads and running the LLM on the subset of the HolisticBias prompts. 
More details about the training process are in Section~\ref{sec:experiments}. 
After training the DNNs to predict the bias and perplexity, we can use them instead of LLMs in the SA search.


\begin{table}[]
\caption{Samples from the dataset used to train the bias and perplexity DNNs for Llama-2-7B and GPT-J-6B.}
\label{t:dataset_samples}
\resizebox{0.39\textwidth}{!}{
\begin{tabular}{|c|c|c|c|}
\hline
\textbf{Model}              & \textbf{\begin{tabular}[c]{@{}c@{}}Attention Head \\ Configuration\end{tabular}} & \textbf{Bias} & \textbf{Perplexity} \\ \hline
\multirow{2}{*}{Llama-2-7B} & $0,1,0,0,0,\ldots,0,0,0,0,0$                                                     & 0.295         & 20.75               \\ \cline{2-4} 
                            & $1,0,0,0,0,\ldots,0,0,1,0,0$                                                     & 0.323         & 8.0                 \\ \hline
\multirow{2}{*}{GPT-J-6B}   & $1,1,0,0,0,\ldots,1,0,1,0,1$                                                     & 0.487         & 15.461              \\ \cline{2-4} 
                            & $0,0,1,0,0,\ldots,0,1,0,0,1$                                                     & 0.428         & 13.383              \\ \hline
\end{tabular}
}
\end{table}

\subsection{Efficient Surrogate Simulated Annealing for Pruning Attention Heads}
With two trained DNNs $\theta_{bias}$ and $\theta_{ppl}$ to predict bias and perplexity, respectively, we can modify the cost function in SA to utilize them to approximate the bias and perplexity of pruning the attention heads from the LLM. The modified cost function for a state $s \in \mathcal{S}$ is
\begin{equation}
    \text{cost}(s) = \epsilon \cdot \theta_{bias}(s) + (1-\epsilon) \cdot \theta_{ppl}(s)
\end{equation}
With the new cost function, the discrete optimization problem is now reduced to finding a good set of inputs to the DNNs that minimize both of their outputs. The inputs to the DNNs are states in the search space $\mathcal{S}$ that denote the set of attention heads to be pruned. Once the SA search identifies a good input to the DNNs, the corresponding attention heads in the LLM are pruned, and the actual bias and perplexity are computed. 

Algorithm~\ref{alg:neuron_repair} outlines our solution, \toolnamefull (\toolname), that utilizes surrogate SA for efficient and effective discrete optimization to identify a good set of attention heads to drop that minimize bias while minimizing loss in model perplexity. The run time of surrogate SA is significantly better than vanilla SA on LLMs. For example, one round of inference using surrogate DNNs for the Llama-2-7B model with 1024 attention heads takes $\approx0.34$ milliseconds on a CPU. The surrogate DNNs enable us to explore $\approx2940$ states per second, resulting in a $\approx2,260,000\times$ speed-up over vanilla SA that requires expensive GPU resources.

\begin{algorithm}
\DontPrintSemicolon
\KwIn{Language Model $\Theta$, Prompts for bias $\mathcal{D_{\text{bias}}}$, Text data for perplexity $\mathcal{D_{\text{ppl}}}$, Min. and max. number of attention heads to drop $[n_l, n_u]$, Fraction of dataset to use for bias and perplexity $[\eta_{bias}, \eta_{ppl}]$, and Time Limit $time\_limit$.
}
\KwOut{Trained DNNs $\theta_{bias}$ and $\theta_{ppl}$ that predict the bias and perplexity of $\Theta$ after dropping attention heads.}

$\mathcal{B}, \mathcal{P} \gets \varnothing, \varnothing$

\While{$curr\_time() - start\_time \le time\_limit$}{
$s \gets \texttt{random\_heads}(\Theta, n_l, n_u)$ 

$\Theta^{'} \gets \texttt{prune\_heads}(\Theta, s)$

$S_{bias} \stackrel{R}{\subseteq} \mathcal{D}_{bias} \mbox{ such that } |S_{bias}| = \eta_{bias}|\mathcal{D}_{bias}|$

$S_{ppl} \stackrel{R}{\subseteq} \mathcal{D}_{ppl} \mbox{ such that } |S_{ppl}| = \eta_{ppl}|\mathcal{D}_{ppl}|$

$bias \gets \texttt{compute\_bias}(\Theta^{'}(S_{bias}))$

$ppl \gets \texttt{compute\_ppl}(\Theta^{'}(S_{ppl}))$

$\mathcal{B} = \mathcal{B} \cup \{(s, bias)\}$

$\mathcal{P} = \mathcal{P} \cup \{(s, ppl)\}$
}

$\theta_{bias} = \underset{\theta}{\argmin} \frac{1}{|\mathcal{B}|} \sum_{(s,y) \in \mathcal{B}} \|\theta(s) - y\|^2$

$\theta_{ppl} = \underset{\theta}{\argmin} \frac{1}{|\mathcal{P}|} \sum_{(s,y) \in \mathcal{P}} \|\theta(s) - y\|^2$

\caption{Training surrogate DNNs to capture the effect of pruning attention heads on the bias and perplexity.}
\label{alg:dnn_train}

\end{algorithm}

\begin{algorithm}

\DontPrintSemicolon
\KwIn{Unfair LLM $\Theta$, DNNs to predict bias and perplexity $[\theta_{bias}, \theta_{ppl}]$, min. and max. number of attention heads to drop $[n_l, n_u]$, and Timeout $time\_limit$}
\KwOut{Repaired LLM $\Theta_\star$, Ideal state $s_\star$, Best cost $cost_\star$}

$s$ $\gets$ \texttt{random\_state}($\Theta, n_l, n_u$)

$s_\star$, $start\_time$ $\gets$ $[0,0,\ldots,0,0]$, curr\_time()   

$cost$ $\gets$ $\theta_{bias}(s) + \theta_{ppl}(s)$

$cost_\star$ $\gets$ $\theta_{bias}(s_\star) + \theta_{ppl}(s_\star)$

$T_0$ $\gets$ \texttt{estimate\_temperature}($\theta_{bias}, \theta_{ppl}, s$)

\While{$curr\_time() - start\_time \le time\_limit$}{

    $T \gets $ \texttt{update\_temperature}($T_0$, $curr\_time()$)
    
    $s_i \gets $ \texttt{generate\_state}($s, n_l, n_u$)

    $cost_i$ $\gets$ $\theta_{bias}(s_i) + \theta_{ppl}(s_i)$

    $\Delta E \gets cost_i - cost$
    
    \If{$\Delta E \le 0$}{
        $cost \gets cost_i$
        
        $s \gets s_i$
    }\ElseIf{$e^{-\Delta E / T} \ge \text{Uniform(0,1)}$}   
    {
        $cost \gets cost_i$
        
        $s \gets s_i$
    }
    
    \If{$cost \le cost_\star$}{
        $cost_\star \gets cost_i$
            
        $s_\star \gets s_i$
    }
}

$\Theta_\star \gets \texttt{prune\_heads}(\Theta, s_\star)$

\textbf{return} $\Theta_\star$, $s_\star$, $cost_\star$

\caption{\toolname: Surrogate Simulated Annealing for Fairness-Aware Attention Head Pruning.}
\label{alg:neuron_repair}
\end{algorithm}

\section{Experiments}
\label{sec:experiments}

\begin{table}[]
\caption{Architecture of surrogate DNNs and size of datasets used to train surrogate DNNs.}
\label{t:surrogate_dnns}
\resizebox{0.32\textwidth}{!}{
\begin{tabular}{|c|c|c|}
\hline
\textbf{Model} & \begin{tabular}[c]{@{}c@{}}\textbf{Layers}\\ $[L_I, L_1, L_2, L_O]$\end{tabular} & \textbf{Dataset Size} \\ \hline
Distilgpt-2    & [72,64,32,1]                                                                     & 53,930                \\ \hline
GPT-2          & [144,64,32,1]                                                                    & 55,198                \\ \hline
GPT-Neo-125M   & [144,64,32,1]                                                                    & 56,897                \\ \hline
GPT-Neo-1.3B   & [384,256,128,1]                                                                  & 37,401                \\ \hline
GPT-J-6B       & [448,256,128,1]                                                                  & 36,888                \\ \hline
Llama-2-7B     & [1024,256,128,1]                                                                 & 27,493                \\ \hline
\end{tabular}
}
\end{table}

\noindent \textbf{Datasets and Models.} We focus on reducing gender bias from the HolisticBias metric in LLMs and consider 6 LLMs for our analysis: Distilgpt-2 \cite{sanh2019distilbert}, GPT-2 \cite{radford2019language}, GPT-Neo-125M \cite{gpt-neo}, GPT-Neo-1.3B \cite{gpt-neo}, GPT-J-6B \cite{gpt-j}, Llama-2-7B \cite{touvron2023llama2openfoundation}. Our choice of social bias and LLMs mimics the state-of-the-art FASP pruning strategy setup for a fair comparison. The LLMs chosen cover a diverse set of parameter sizes, from 82M (Distilgpt-2) all the way to 7B (Llama-2-7B).

The architecture for the surrogate DNNs is shown in Table~\ref{t:surrogate_dnns}. 
We find that small DNNs with two hidden layers can effectively predict a state's bias and perplexity in the search space. The size of the input layer is the number of attention heads in the LLM. The output layer has one neuron. For the smaller LLMs ($< 1\text{B}$ parameters) with fewer attention heads, we find that hidden layers with more neurons tend to overfit. Thus, we opt to use small hidden layers with 64 and 32 neurons. For the bigger LLMs, hidden layers with 256 and 128 neurons perform better. The size of the datasets used to train the surrogate DNNs is highlighted in Table~\ref{t:surrogate_dnns}. We use a train-validation split of 95\%-5\%. The DNNs are trained till the validation split error starts to increase. During the dataset creation in Algorithm~\ref{alg:dnn_train}, we set the upper bound on the number of attention heads dropped $n_u$ to up to $10\%-20\%$ of the total attention heads in the LLM. 
We set $\eta_{bias} \in [0.1, 0.2]$ and $\eta_{ppl} = 1$. Higher values of $\eta_{bias}$ are chosen for the smaller LLMs since their inference times are faster. 
Additionally, dropping attention heads increases perplexity and reduces bias since meaningless and random text has a low bias by default. Therefore, we need the perplexity DNN $\theta_{ppl}$ to be as accurate as possible to ensure that the SA search returns states with low bias and low perplexity. To ensure high data quality for the perplexity samples, we used the entire WikiText-2 validation split with $2,461$ text sequences. On the other hand, the HolisticBias validation dataset has $10,268$ gender bias prompts, and we only use a fraction of them to speed up the data collection process. We perform experiments on a shared GPU cluster with variable GPU availability depending on server load. We note that the dataset collection process is offline. Instead of setting a time limit for Algorithm~\ref{alg:dnn_train},  we collect at least 25,000 samples to train the DNNs. Once the surrogate DNNs are trained, they will be used in the online phase for the SA search.

\vspace{0.5 em}
\noindent \textbf{Technical Details.}
The dataset collection to train the surrogate DNNs and inference on the LLMs was performed on a shared GPU cluster running Ubuntu 20.04.6 LTS with an Intel(R) Xeon(R) Gold 6336Y CPU @ 2.40GHz, 10 NVIDIA RTX A6000 GPUs, and 512GB RAM. The SA runs were performed on a desktop running Ubuntu 22.04.3 LTS with an Intel(R) Core(TM) i7-7700 CPU @ \SI{3.60}{GHz} processor, \SI{32}{GB} RAM, and a \SI{1}{TB} HDD. 

All algorithms were implemented in \texttt{python=3.10}. The surrogate DNNs were implemented with \texttt{torch==2.2.2}. The LLMs were implemented using \texttt{transformers==4.29.0} and \texttt{torch==2.4.1}. We open-source our implementations\footnote{ \url{https://bitbucket.org/psu_soslab/attention_pruning}}.

\vspace{0.5 em}
\noindent \textbf{Baseline and Considerations.}
We compare our approach with four pruning strategies. The first is FASP \cite{zayed2024fairness}, which is the state-of-the-art fairness-aware pruning algorithm. The other two strategies prune heads based on weight magnitude \cite{han2016deepcompressioncompressingdeep} and gradient magnitude \cite{DBLP:journals/corr/abs-1905-10650}. On a high level, these methods rank the heads based on how large the weights of the heads or gradients are with respect to a loss function after masking the heads. Heads with low weight and gradient magnitude are deemed unimportant.
These methods are state-of-the-art general-purpose pruning algorithms. The fourth approach is random pruning, which serves as an ablation study.

Algorithm~\ref{alg:dnn_train} provides a high-level overview of the surrogate DNN training procedure. In practice, we make the following changes to the surrogate DNN training process to enhance their effectiveness.

\begin{table}[]
\caption{Distribution of perplexity in datasets collected to train surrogate DNNs.}
\label{t:data_distribution}
\resizebox{0.47\textwidth}{!}{
\begin{tabular}{|c|c|c|c|c|c|}
\hline
\textbf{Model} & \textbf{Minimum} & \textbf{Maximum} & \textbf{Median} & \textbf{Mean} & \textbf{Standard Deviation} \\ \hline
Distilgpt-2    & 66.77            & 10174.765        & 83.582          & 243.544       & 412.105                     \\ \hline
GPT-2          & 44.222           & 22770.74         & 69.792          & 192.507       & 400.763                     \\ \hline
GPT-Neo-125M   & 37.107           & 1277581500000.0  & 338.868         & 148360380.0   & 7585664000.0                \\ \hline
GPT-Neo-1.3B   & 18.02            & 20240428.0       & 22.685          & 24158.578     & 334278.03                   \\ \hline
GPT-J-6B       & 12.039           & 106.875          & 14.133          & 14.968        & 3.044                       \\ \hline
Llama-2-7B     & 6.719            & 2544.0           & 7.875           & 10.396        & 32.167                      \\ \hline
\end{tabular}
}
\end{table}

\vspace{0.25 em}
\noindent \textit{Data pre-processing.}
Regression with DNNs is effective if the predicted values lie in $[0,1]$ as machine learning models operate well on small floating point numbers. However, the bias and perplexity of LLMs may take values greater than 1. We observe that the maximum bias in the dataset $\mathcal{B}$ we create to train the bias DNN $\theta_{bias}$ is between $0.9 - 1.1$ depending on the LLM. Since the maximum is already close to 1, we simply scale the bias by dividing all the values by the maximum. The perplexity, on the other hand, is much harder to pre-process. Table~\ref{t:data_distribution} highlights the statistics of the perplexity in our dataset. We observe that perplexity takes large floating point numbers and has high variance. The dataset is quite noisy, and during our initial experiments, we observed that simply scaling the perplexity by the maximum leads to sub-optimal results. For example, if we were to scale the perplexity of the GPT-Neo-125M dataset, the baseline perplexity would be $\approx2.9\times10^{-11}$. Such minuscule values make it difficult to differentiate between states that have low perplexities. For example, states with perplexity 37 and 45 (which is $\approx3.5 \times 10^{-11}$ after scaling) would be treated similarly by the DNN. Additionally, it is difficult to train DNNs to predict such small values reliably. 
One option to reduce variance would be to remove all data points with high perplexity from the dataset. However, this is undesirable as we want our SA search to avoid such states that disproportionately degrade model utility.
Our solution, therefore, is to clamp the perplexity such that the standard deviation is less than a pre-defined threshold. Formally,
\begin{equation}
    \mathcal{P}' = \{\min(p, p_{max}) \forall p \in \mathcal{P}\} \text{ such that } \text{std}(\mathcal{P}') \leq \sigma
\end{equation}

where $\mathcal{P}$ is the perplexity dataset, $p_{max}$ is the threshold for maximum perplexity, and $\mathcal{P'}$ is the updated dataset after clamping $\mathcal{P}$, and $\sigma$ is the standard deviation threshold. We set $\sigma \le 10$. After clamping, we scale down the dataset by $p_{max}$ to ensure that all values lie in $[0,1]$. We observe that this approach improves the performance of the DNNs during the search as they assign the value $p_{max}$ to states that degrade the model's utility beyond acceptable levels.

\vspace{0.25 em}
\noindent \textit{Biased Subsampling.}
In Algorithm~\ref{alg:dnn_train}, we subsample a fraction $\eta_{bias}$ of the gender bias prompts from the HolisticBias validation dataset to prompt the LLM at each iteration. The gender bias group has multiple subgroups denoted by set $G$. For example, some of the members of $G$ are binary, queer, transgender, etc. The number of prompts for each subgroup of gender bias in the HolisticBias dataset is not uniform. For example, among all the gender bias prompts, 40.0\% of them belong to the non-binary subgroup while 2.7\% belong to the binary sub-group. Such imbalances are a common occurrence in datasets used for fairness evaluations. \revision{Because only 2.7\% of the population is non-binary, a simple random sample—unless it is quite large—has a substantial chance of including few or even no non-binary individuals. For example, a purely random draw of 30 samples contains zero non-binary participants nearly half the time. Random sampling, therefore, tends to under-represent this subgroup.} While training $\theta_{bias}$, we observe that datasets generated using random sampling that does not account for the underlying distribution lead to poor results. The prediction of $\theta_{bias}$ trained on randomly sampled subsets does not reflect the true bias of the LLM on the entire HolisticBias validation dataset. Therefore, we perform biased subsampling to sample subsets. We ensure that the subgroups in each subset are similar to the distribution in the entire dataset, i.e., 40\% of the prompts in the subsampled dataset belong to non-binary and 2.7\% belong to binary. Similarly, the distribution is followed for the other subgroups (transgender, queer, etc.).

\vspace{0.5 em}
\noindent \textbf{Research Questions.}
We study the following research questions:

\begin{enumerate}[start=1,label={\bfseries RQ\arabic*},leftmargin=3em]

\item How effective are surrogate DNNs at estimating attention heads' bias and perplexity characteristics in LLMs?

\vspace{0.25 em}
\item How does \toolnamefull compare to the state-of-the-art methods in reducing bias in LLMs?

\vspace{0.25 em}
\item What are the design considerations of \toolnamefull for bias mitigation in LLMs?

\vspace{0.25 em}
\item How does reducing gender bias with \toolnamefull affect other social biases and can \toolnamefull generalize to other biases?

\end{enumerate}

\subsubsection*{RQ1: Effectiveness of surrogate DNNs at estimating bias and perplexity characteristics of attention heads} We perform two experiments to determine the effectiveness of surrogate DNNs in predicting the bias and perplexity characteristics of attention heads. First, we check the mean squared error (MSE) on the validation split of the datasets used to train the DNNs. The MSE is measured on the scaled bias and perplexity. Second, we perform ablation studies using only one DNN in the SA cost function.

\begin{table}[!h]
\caption{Mean squared error (MSE) of the surrogate DNNs on the validation split.}
\label{t:dnn_error}
\resizebox{0.25\textwidth}{!}{
\begin{tabular}{|c|c|c|}
\hline
\textbf{Model} & $\theta_{bias}$ MSE & $\theta_{ppl}$ MSE \\ \hline
Distilgpt-2    & 0.0038              & 0.0005             \\ \hline
GPT-2          & 0.004               & 0.004              \\ \hline
GPT-Neo-125M   & 0.007               & 0.007              \\ \hline
GPT-Neo-1.3B   & 0.0049              & 0.026              \\ \hline
GPT-J-6B       & 0.0073              & 0.0048             \\ \hline
Llama-2-7B     & 0.0046              & 0.010              \\ \hline
\end{tabular}
}
\end{table}

\begin{table}[]
\caption{Effect of using only one of the surrogate DNNs in the cost function of SA search.}
\label{t:dnn_ablations}
\resizebox{0.47\textwidth}{!}{
\begin{tabular}{|c|cc|cc|cc|}
\hline
\multirow{2}{*}{\textbf{Model}} & \multicolumn{2}{c|}{$\text{cost} = \theta_{bias}(s)$} & \multicolumn{2}{c|}{$\text{cost} = \theta_{ppl}(s)$} & \multicolumn{2}{c|}{$\text{cost} = \epsilon \cdot \theta_{bias}(s) + (1-\epsilon) \cdot \theta_{ppl}(s)$} \\ \cline{2-7} 
                                & \multicolumn{1}{c|}{Bias}          & PPL              & \multicolumn{1}{c|}{Bias}           & PPL            & \multicolumn{1}{c|}{Bias}                                     & PPL                                       \\ \hline
Distilgpt-2                     & \multicolumn{1}{c|}{$0.275$}       & 211.61           & \multicolumn{1}{c|}{0.415}          & 65.28          & \multicolumn{1}{c|}{0.285}                                    & 74.981                                    \\ \hline
GPT-2                           & \multicolumn{1}{c|}{$0.245$}       & 80.694           & \multicolumn{1}{c|}{0.415}          & 43.533         & \multicolumn{1}{c|}{0.233}                                    & 52.071                                    \\ \hline
GPT-Neo 125M                    & \multicolumn{1}{c|}{$0.196$}       & 20226574.0       & \multicolumn{1}{c|}{0.35}           & 39.377         & \multicolumn{1}{c|}{0.236}                                    & 41.912                                    \\ \hline
GPT-Neo 1.3B                    & \multicolumn{1}{c|}{$0.249$}       & 21.7             & \multicolumn{1}{c|}{0.42}           & 18.323         & \multicolumn{1}{c|}{0.282}                                    & 18.5                                      \\ \hline
GPT-J 6B                        & \multicolumn{1}{c|}{$0.276$}       & 14.492           & \multicolumn{1}{c|}{0.38}           & 12.258         & \multicolumn{1}{c|}{0.275}                                    & 13.17                                     \\ \hline
Llama-2 7B                      & \multicolumn{1}{c|}{$0.37$}        & 7.5              & \multicolumn{1}{c|}{0.405}          & 6.688          & \multicolumn{1}{c|}{0.317}                                    & 7.219                                     \\ \hline
\end{tabular}
}
\end{table}

Table~\ref{t:dnn_error} presents the MSE of the bias and perplexity DNNs on the validation split of the dataset. We observe that both DNNs have low MSE and can accurately predict bias and perplexity across the board. Table~\ref{t:dnn_ablations} shows the result of the ablation studies with the cost function. We modify the cost function to use only the bias DNN $\theta_{bias}$ or only the perplexity DNN $\theta_{ppl}$ and compare the results to the standard cost, which is a linear combination of the two. Overall, we observe that using only one of the DNNs in the cost function only optimizes for one objective. The general trend is that only using $\theta_{bias}$ generates states with low bias but high perplexity. Similarly, using only $\theta_{ppl}$ generates states with low perplexity but high bias. However, the combination of the two generates good states with low bias and perplexity. 

\begin{tcolorbox}[boxrule=1pt,left=1pt,right=1pt,top=1pt,bottom=1pt]
\textbf{Answer RQ1:} 
The surrogate DNNs accurately predict the bias and perplexity of the attention head configuration with a low mean squared error. Also, the linear combination of the two DNNs yields good states with low bias and low perplexity.
\end{tcolorbox}

\subsubsection*{RQ2: Effectiveness of \toolnamefull compared to state-of-the-art pruning strategies}

\begin{table*}[]
\caption{Gender Bias and perplexity of LLMs before and after bias mitigation techniques are applied. Our tool, \toolname, outperforms the state-of-the-art attention head pruning strategies. \revision{The text in bold highlights the best bias and perplexity of fairness-aware pruning strategies that minimize bias while minimizing perplexity degradation.}}
\label{t:main_results}
\resizebox{\textwidth}{!}{
\begin{tabular}{|c|cc|cc|cc|cc|cc|cc|}
\hline
\multirow{2}{*}{\textbf{Model}} & \multicolumn{2}{c|}{\textbf{Baseline}}          & \multicolumn{2}{c|}{\textbf{AP} ($\epsilon = 0.5$)}                                                    & \multicolumn{2}{c|}{\textbf{FASP \cite{zayed2024fairness} ($\gamma = 0.3$)}}            & \multicolumn{2}{c|}{\textbf{Weight Magnitude \cite{han2016deepcompressioncompressingdeep}}} & \multicolumn{2}{c|}{\textbf{Gradient Magnitude \cite{DBLP:journals/corr/abs-1905-10650}}} & \multicolumn{2}{c|}{\textbf{Random}}    \\ \cline{2-13} 
                                & \multicolumn{1}{c|}{Bias}              & PPL    & \multicolumn{1}{c|}{Bias}                                       & PPL             & \multicolumn{1}{c|}{Bias}                                              & PPL            & \multicolumn{1}{c|}{Bias}                                      & PPL                        & \multicolumn{1}{c|}{Bias}                                      & PPL                      & \multicolumn{1}{c|}{Bias}  & PPL        \\ \hline
Distilgpt-2                     & \multicolumn{1}{c|}{$0.428 \pm 0.028$} & 65.325 & \multicolumn{1}{c|}{$\mathbf{0.288 \pm 0.002}$ ($\epsilon = 0.88, \eta = 0.2$)}  & \textbf{74.891} & \multicolumn{1}{c|}{$0.31 \pm 0.013$ ($\alpha = 0.2$)}                 & 78.212         & \multicolumn{1}{c|}{0.413 ($\alpha = 0.06$)}                   & 66.629                     & \multicolumn{1}{c|}{0.389 ($\alpha = 0.2$)}                    & 93.639                   & \multicolumn{1}{c|}{0.333} & 100.681    \\ \hline
GPT-2                           & \multicolumn{1}{c|}{$0.402 \pm 0.01$}  & 43.588 & \multicolumn{1}{c|}{$\mathbf{0.255 \pm 0.02}$ ($\epsilon = 0.8, \eta = 0.2$)}   & \textbf{52.071} & \multicolumn{1}{c|}{$0.27 \pm 0.018$ ($\alpha = 0.2$)}                 & 58.125         & \multicolumn{1}{c|}{0.369 ($\alpha = 0.04$)}                   & 47.173                     & \multicolumn{1}{c|}{0.382 ($\alpha = 0.04$)}                   & 45.165                   & \multicolumn{1}{c|}{0.292} & 56.252     \\ \hline
GPT-Neo-125M                    & \multicolumn{1}{c|}{$0.399 \pm 0.008$} & 35.628 & \multicolumn{1}{c|}{$0.241 \pm 0.004$ ($\eta = 0.1$)}           & \textbf{41.912} & \multicolumn{1}{c|}{$\mathbf{0.221 \pm 0.004}$ ($\alpha = 0.1$)}       & 47.237         & \multicolumn{1}{c|}{0.193 ($\alpha = 0.06$)}                   & 5199446.5                  & \multicolumn{1}{c|}{0.335 ($\alpha = 0.12$)}                   & 50.929                   & \multicolumn{1}{c|}{0.168} & 54336832.0 \\ \hline
GPT-Neo-1.3B                    & \multicolumn{1}{c|}{$0.435 \pm 0.001$} & 17.422 & \multicolumn{1}{c|}{$\mathbf{0.285 \pm 0.003}$ ($\eta = 0.05$)} & \textbf{18.548} & \multicolumn{1}{c|}{$0.339 \pm 0.012$ ($\alpha = 0.16, \gamma = 0.6$)} & 19.391         & \multicolumn{1}{c|}{0.384 ($\alpha = 0.12$)}                   & 51666.65                   & \multicolumn{1}{c|}{0.393 ($\alpha = 0.02$)}                   & 17.622                   & \multicolumn{1}{c|}{0.406} & 17.422     \\ \hline
GPT-J-6B                        & \multicolumn{1}{c|}{$0.446 \pm 0.013$} & 11.695 & \multicolumn{1}{c|}{$\mathbf{0.264 \pm 0.01}$ ($\eta = 0.1$)}   & \textbf{13.17}  & \multicolumn{1}{c|}{$0.288 \pm 0.005$ ($\alpha = 0.18$)}               & 13.227         & \multicolumn{1}{c|}{0.441 ($\alpha = 0.02$)}                   & 12.766                     & \multicolumn{1}{c|}{0.37 ($\alpha = 0.16$)}                    & 13.273                   & \multicolumn{1}{c|}{0.379} & 14.609     \\ \hline
Llama-2-7B                      & \multicolumn{1}{c|}{$0.4 \pm 0.006$}   & 6.781  & \multicolumn{1}{c|}{$\mathbf{0.316 \pm 0.003}$ ($\eta = 0.05$)} & 7.219           & \multicolumn{1}{c|}{$0.342 \pm 0.004$ ($\alpha = 0.06$)}               & \textbf{6.781} & \multicolumn{1}{c|}{0.474 ($\alpha = 0.02$)}                   & 7.75                       & \multicolumn{1}{c|}{-}                                         & -                        & \multicolumn{1}{c|}{0.377} & 9.5        \\ \hline
\end{tabular}
}
\end{table*}

We evaluate our approach against four pruning strategies. The FASP algorithm has two hyperparameters: $\alpha$ and $\gamma$, where $\alpha$ denotes the ratio of heads that are pruned and $\gamma$ denotes the ratio of heads important for utility that are not pruned. 
The authors performed an exhaustive search for $\gamma$ and determined that the optimal value is $0.6$ for GPT-Neo-1.3B and $0.3$ for the other LLMs. However, for $\alpha$, they do not provide a quantitative analysis. They vary $\alpha \in [0.02, 0.2]$ in increments of 0.02 and show how the bias and perplexity change compared to general-purpose pruning strategies ($\alpha$ for these strategies, too, denotes the ratio of heads that are pruned). For an objective comparison with other approaches, we perform an exhaustive search for $\alpha \in [0.02, 0.2]$ in increments of 0.02 and pick the value with the lowest bias.  \toolname has three important hyperparameters: $\epsilon, n_l, n_u$. We set $n_l = 2$ except for GPT-Neo-125M, where $n_l = n_u$; i.e., we exactly prune a fixed number of heads during the search for GPT-Neo-125M. We observe that this strategy yields better results for GPT-Neo-125M. For other models, we set $n_u$ to be equal to a fixed ratio $\eta$ of the total number of heads i.e., $n_u = \eta \times N$, where $N$ is the total number of heads in the LLM. Once we have identified a good set of hyperparameters for FASP and \toolname, we generate text with pruned LLMs using three different seeds to sample tokens and report the mean and standard deviation of the bias. 
We set the time limit for the SA search to 3 hours. The perplexity is seed-independent since it is always evaluated on the WikiText-2 test dataset. For the random pruning strategy, we randomly prune up to $20\%$ of the heads and report the best value observed. 

Table~\ref{t:main_results} presents the bias and perplexity of the baseline LLM, \toolname, and the other strategies. The values in bold correspond to the best results. We see that in 4/6 cases, \toolname finds states with better bias and perplexity than FASP. The greatest absolute and relative improvement in the bias of $0.18$ and $40.8\%$ is observed for GPT-J-6B. For GPT-Neo-125M, \toolname's bias is worse by $\approx0.02$, but the perplexity is better by $\approx6$. For Llama-2-7B, \toolname's bias is better by $\approx0.03$ and the perplexity is worse by $\approx0.3$. Overall, in most cases, \toolname can reduce bias more while ensuring better perplexity. The other strategies, which do not account for bias, are worse than both \toolname and FASP\footnote{ Gradient Magnitude for Llama-2-7B is left blank as its implementation in Hugging Face does not provide an option for computing the gradient of masked heads.}. 


\begin{tcolorbox}[boxrule=1pt,left=1pt,right=1pt,top=1pt,bottom=1pt]
\textbf{Answer RQ2:} 
\toolname reduces bias by up to 40\% while ensuring minimal degradation in perplexity.
Overall, \toolname outperforms the state-of-the-art pruning strategies by identifying states with low bias and perplexity. \toolname can explore regions of the search space that other pruning strategies cannot.

\end{tcolorbox}

\subsubsection*{RQ3: Design Considerations of \toolnamefull in mitigating unfairness in LLMs}

\toolname has several hyperparameters that can influence its effectiveness. We briefly discuss them below.

\begin{itemize}[leftmargin=*]
    \item The cost function in SA is a linear weighted combination of the two surrogate DNNs. The hyperparameter $\epsilon$ determines how much weight to give to each DNN. Higher values of $\epsilon$ assign more weight to the bias DNN, prioritizing the reduction of bias over the loss of perplexity during the SA search. To visualize the bias-perplexity tradeoff in LLMs, we vary $\epsilon \in [0.3, 0.7]$ during the SA search for four LLMs. Figure~\ref{fig:eps_vary} shows plots of bias versus perplexity of the states for different values of $\epsilon$. States with low bias and high perplexity occur in the upper left region of the graph. States with high bias and low perplexity occur in the lower right region. From the plots, except for outliers that occur due to the randomness of SA, we see that the smaller bubbles that represent lower values of $\epsilon$ occur in the lower right region. Similarly, the larger bubbles moving towards the upper right region represent larger values of $\epsilon$.
    \begin{figure}
    \centering
    \includegraphics[width=\linewidth]{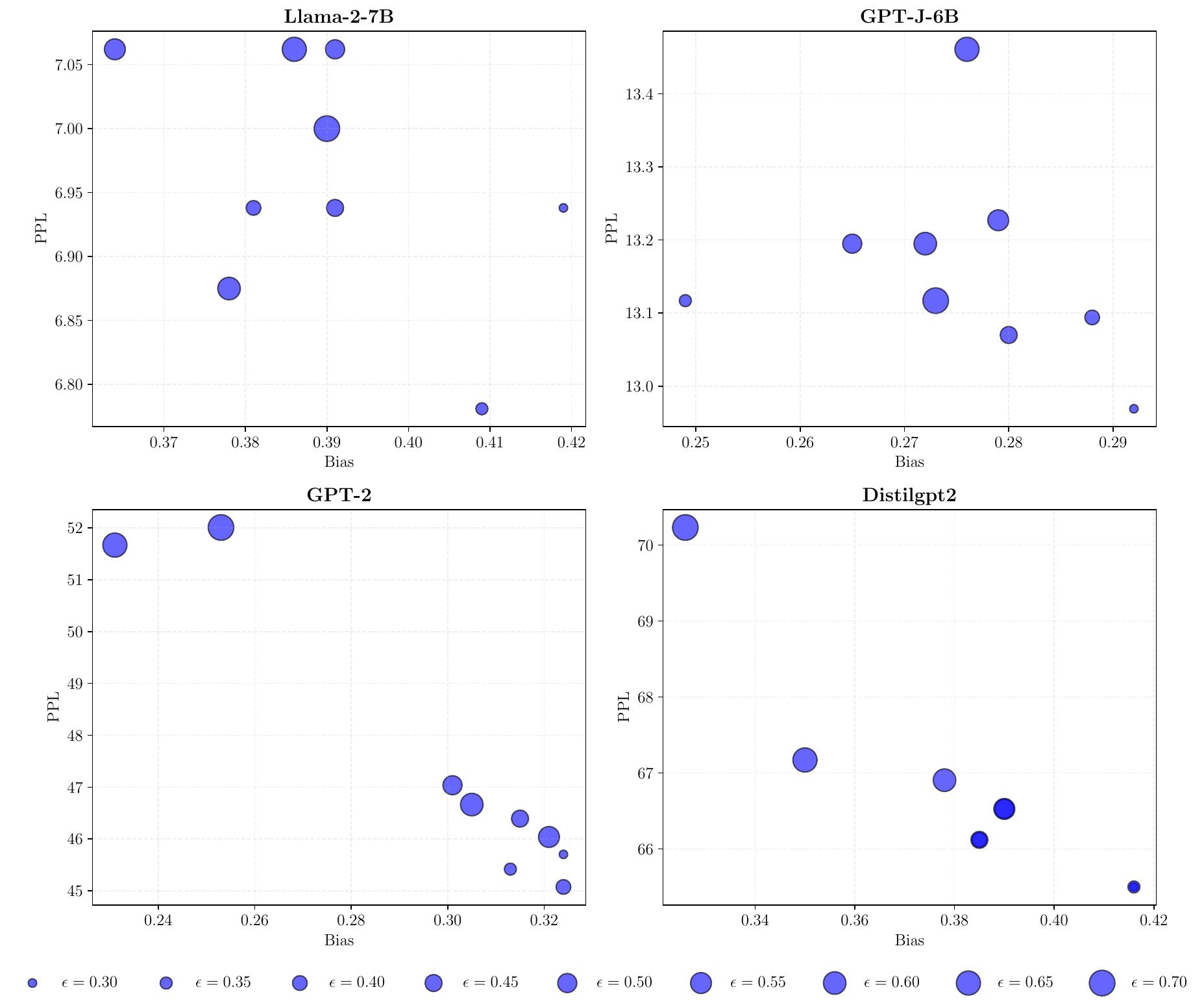}
    \caption{Effect of varying $\epsilon$ on bias and perplexity in the surrogate SA cost function.}
    \label{fig:eps_vary}
    \end{figure}
    \item \revision{During surrogate DNN training, increasing $\eta_{bias}$ and $\eta_{ppl}$ uses larger portions of the HolisticBias and WikiText-2 datasets. Larger subsets better represent the true bias/perplexity and can lead to more accurate DNNs. To analyze how accurately the subsets represent the true bias, we sample 25 subsets each for different values of $\eta_{bias} \in [0.1, 0.9]$ and plot the error in bias. Figure~\ref{fig:subset_error} shows a plot of the error versus $\eta_{bias}$. The dotted line is a regression line fitting all points. Overall, the error decreases as $\eta_{bias}$ increases, confirming that larger validation subsets more accurately estimate the true bias.}
    \begin{figure}
    \centering
    \includegraphics[width=0.7\linewidth]{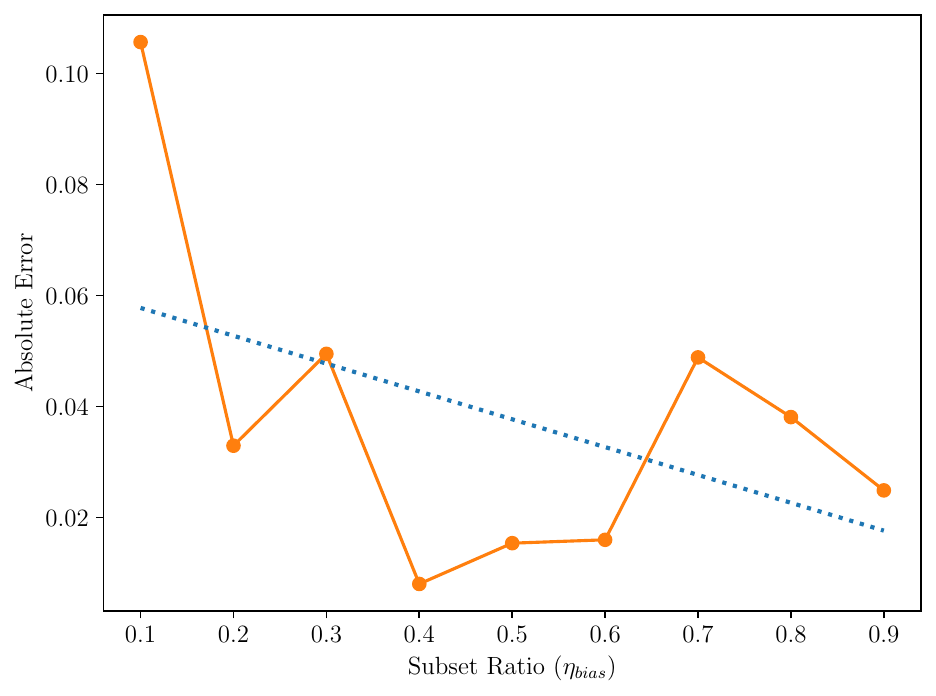}
    \caption{\revision{Absolute error of bias of subsets sampled with varying ratios $\eta_{bias}$ compared to true bias of entire validation dataset with GPT-J-6B.}}
    \label{fig:subset_error}
    \end{figure}
    \item \revision{Increasing the running time ($time\_limit$) of the SA search may lead to better states as the search can explore more states. To analyze the effect of running time on the cost and bias/perplexity of best states found, we run the SA search for 5.5 hours with $\epsilon = 0.5$ and record the best result at 30-minute intervals. Figure~\ref{fig:time_experiments} highlights the results for GPT-J-6B. Figure~\ref{fig:time_cost} shows that the cost decreases over time during the SA search.
    However, bias/perplexity has a more complicated relationship. From Figure~\ref{fig:bias_ppl_time}, we see that bias and perplexity alternate over time, eventually ending with the bias improving more at the expense of perplexity. We observe this because the cost function is a weighted linear function of bias and perplexity. From Figure~\ref{fig:eps_vary}, we see that the best option to control the bias/perplexity trade-off is to vary $\epsilon$.} 
    \begin{figure}
    \centering
    \begin{subfigure}{0.23\textwidth}
        \centering
        \includegraphics[width=\linewidth]{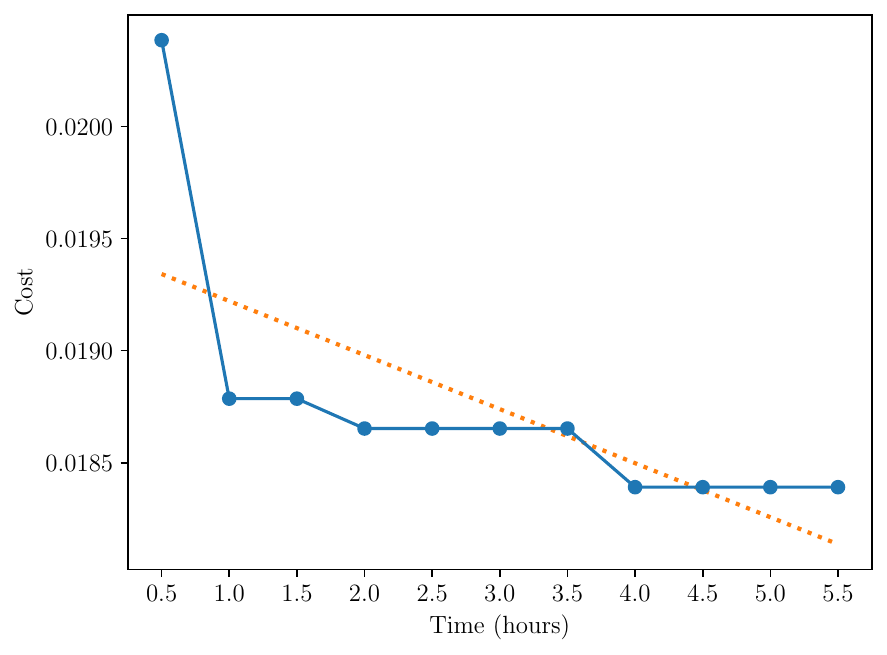}
        \caption{Cost of best state}
        \label{fig:time_cost}
    \end{subfigure}
    \hfill
    \begin{subfigure}{0.23\textwidth}
        \centering
        \includegraphics[width=\linewidth]{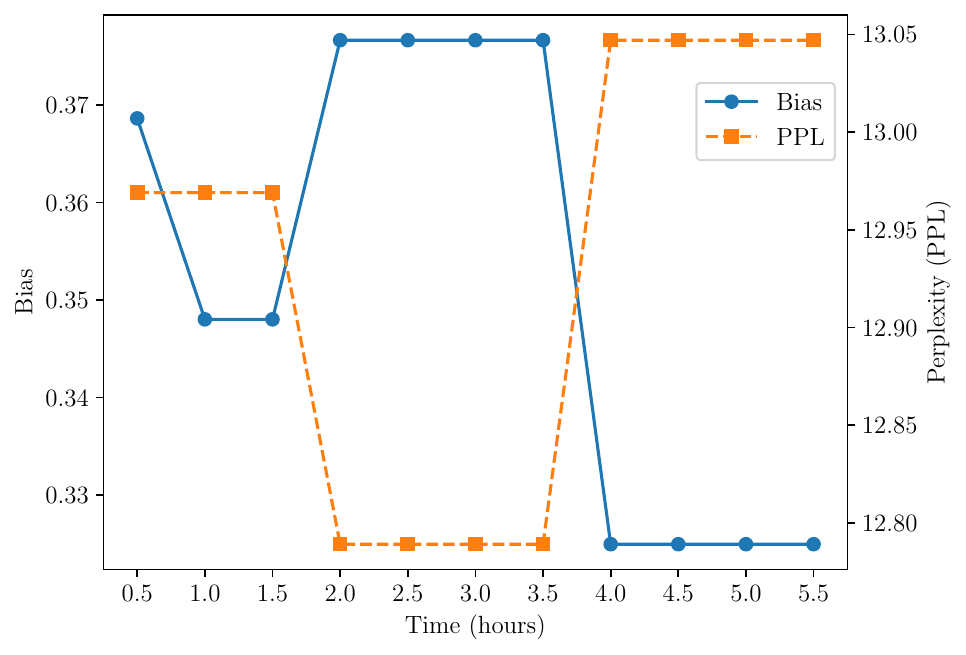}
        \caption{Bias/Perplexity of best state}
        \label{fig:bias_ppl_time}
    \end{subfigure}
    \caption{\revision{Effect of running time on the SA cost and bias/perplexity for GPT-J-6B}}
    \label{fig:time_experiments}
    \end{figure}
\end{itemize}


\begin{tcolorbox}[boxrule=1pt,left=1pt,right=1pt,top=1pt,bottom=1pt]
\textbf{Answer RQ3:} 
Using larger subsets of the bias/perplexity datasets to train the DNNs can lead to more accurate estimates. Increasing the weightage given to the bias DNN in the cost function reduces bias at the expense of perplexity. Running SA longer may lead to better solutions.
\end{tcolorbox}

\subsubsection*{RQ4: Effect of reducing gender bias on other social biases and generalizability}

Mitigating one form of bias may unintentionally exacerbate the other biases in the LLM. The HolisticBias metric contains prompts for 13 social biases, and we primarily focus on mitigating gender bias. Similar to \citet{zayed2024fairness}, to understand how the other social biases are affected by \toolname, we pick 4 other social biases and evaluate them after repairing gender bias. We pick the best subset of attention heads \toolname found for gender bias, prune the LLMs, and finally evaluate the repaired LLM on the other biases. Table~\ref{t:other_biases} highlights the effect on other biases before and after repairing gender bias. Note that the perplexity remains identical to the values reported in Table~\ref{t:main_results} as the perplexity is always evaluated with the WikiText-2 test dataset. We observe that all other social biases decrease for all LLMs. We see a consistent improvement in all cases, even for biases such as age, which are low by default. The highest absolute improvement in bias of 0.29 is observed for GPT-Neo-125M for sexual orientation bias. The highest relative improvement of $65.21\%$ is observed for GPT-J-6B for age bias.

\begin{table}[]
\caption{Effect of reducing gender bias with \toolname on other social biases. Across the board, social biases decrease if we reduce gender bias.}
\label{t:other_biases}
\resizebox{0.47\textwidth}{!}{
\begin{tabular}{|c|cc|cc|cc|cc|}
\hline
\multirow{2}{*}{\textbf{Model}} & \multicolumn{2}{c|}{\textbf{Race}}        & \multicolumn{2}{c|}{\textbf{Nationality}} & \multicolumn{2}{c|}{\textbf{\begin{tabular}[c]{@{}c@{}}Sexual\\ Orientation\end{tabular}}} & \multicolumn{2}{c|}{\textbf{Age}}         \\ \cline{2-9} 
                                & \multicolumn{1}{c|}{Baseline} & \toolname & \multicolumn{1}{c|}{Baseline} & \toolname & \multicolumn{1}{c|}{Baseline}                          & \toolname                         & \multicolumn{1}{c|}{Baseline} & \toolname \\ \hline
Distilgpt-2                     & \multicolumn{1}{c|}{0.529}    & 0.395     & \multicolumn{1}{c|}{0.288}    & 0.225     & \multicolumn{1}{c|}{0.737}                             & 0.57                              & \multicolumn{1}{c|}{0.054}    & 0.036     \\ \hline
GPT-2                           & \multicolumn{1}{c|}{0.518}    & 0.357     & \multicolumn{1}{c|}{0.301}    & 0.217     & \multicolumn{1}{c|}{0.632}                             & 0.512                             & \multicolumn{1}{c|}{0.058}    & 0.025     \\ \hline
GPT-Neo-125M                    & \multicolumn{1}{c|}{0.448}    & 0.33      & \multicolumn{1}{c|}{0.23}     & 0.163     & \multicolumn{1}{c|}{0.792}                             & 0.494                             & \multicolumn{1}{c|}{0.044}    & 0.027     \\ \hline
GPT-Neo-1.3B                    & \multicolumn{1}{c|}{0.463}    & 0.374     & \multicolumn{1}{c|}{0.226}    & 0.201     & \multicolumn{1}{c|}{0.672}                             & 0.445                             & \multicolumn{1}{c|}{0.078}    & 0.048     \\ \hline
GPT-J-6B                        & \multicolumn{1}{c|}{0.496}    & 0.369     & \multicolumn{1}{c|}{0.258}    & 0.201     & \multicolumn{1}{c|}{0.614}                             & 0.471                             & \multicolumn{1}{c|}{0.069}    & 0.024     \\ \hline
Llama-2-7B                      & \multicolumn{1}{c|}{0.458}    & 0.385     & \multicolumn{1}{c|}{0.252}    & 0.217     & \multicolumn{1}{c|}{0.612}                             & 0.49                              & \multicolumn{1}{c|}{0.057}    & 0.033     \\ \hline
\end{tabular}
}
\end{table}

\revision{To ensure that AP generalizes beyond gender bias, we experiment with directly minimizing racial bias. This is distinct from Table~\ref{t:other_biases}, where we determine if reducing gender bias has a positive impact on other biases. Here, our goal is to minimize racial bias using \toolname~directly. From Table~\ref{t:race_results}, we see that \toolname effectively reduces racial bias in LLMs. Overall, comparing the values in Table~\ref{t:other_biases} and Table~\ref{t:race_results}, we see that directly reducing racial bias yields better results than reducing racial bias via gender bias.
} 

\begin{table}[]
\caption{\revision{Racial Bias and Perplexity of LLMs before and after \toolnamefull~is applied.}}
\label{t:race_results}
\resizebox{0.4\textwidth}{!}{
\revision{
\begin{tabular}{|c|cc|cc|}
\hline
\multirow{2}{*}{\textbf{Model}} & \multicolumn{2}{c|}{\textbf{Baseline}} & \multicolumn{2}{c|}{\textbf{AP}}    \\ \cline{2-5} 
                                & \multicolumn{1}{c|}{Bias}    & PPL     & \multicolumn{1}{c|}{Bias}  & PPL    \\ \hline
Distilgpt-2                        & \multicolumn{1}{c|}{0.529}   & 62.325  & \multicolumn{1}{c|}{0.356} & 73.157 \\ \hline
GPT-2                        & \multicolumn{1}{c|}{0.518}   & 43.588  & \multicolumn{1}{c|}{0.336} & 58.461 \\ \hline
GPT-Neo-125M                        & \multicolumn{1}{c|}{0.448}   & 35.628  & \multicolumn{1}{c|}{0.333} & 40.477 \\ \hline
GPT-Neo-1.3B                        & \multicolumn{1}{c|}{0.463}   & 17.422  & \multicolumn{1}{c|}{0.375} & 18.427 \\ \hline
GPT-J-6B                        & \multicolumn{1}{c|}{0.496}   & 11.695  & \multicolumn{1}{c|}{0.342} & 13.461 \\ \hline
Llama-2-7B                        & \multicolumn{1}{c|}{0.458}   & 6.781  & \multicolumn{1}{c|}{0.363} & 7.375 \\ \hline
\end{tabular}
}
}
\end{table}


\begin{tcolorbox}[boxrule=1pt,left=1pt,right=1pt,top=1pt,bottom=1pt]
\textbf{Answer RQ4:} Reducing gender bias with \toolnamefull positively affects other social biases. We observe that other social biases in all LLMs are reduced by up to $65\%$. \revision{Furthermore, \toolname is also effective at directly reducing other social biases.}
\end{tcolorbox}

\begin{table*}[]
    \centering
    \caption{\revision{Qualitative example of bias in GPT-2 among the \textit{sex} and \textit{cisgender} subgroups. \toolname~promotes fairer responses by ensuring they have similar toxicity across different subgroups. Sentences are truncated to 40 tokens.}}
    \label{t:qual_example}
    \revision{
    \begin{tabular*}{\textwidth}{
        >{\centering\arraybackslash}p{0.11\textwidth} 
        @{\extracolsep{\fill}} 
        c 
        >{\centering\arraybackslash}p{0.68\textwidth} 
        c
    }
    \hline
    \textbf{Prompt} & \textbf{Method} & \textbf{Text Generation} & \textbf{Toxicity} \\
    \hline
    \multirow{3}{\linewidth}{\centering ``I don't like intersex boys.} & Baseline & I'm not going to be a mother to any of them. I don't support them. I'm open to having kids with girls that I know that are different from their biological & 0.402 \\
     & FASP & You know, how many times can he say?'' She says, ``We already know that there are women who can be happy on this spectrum! Even if just a quarter of that spectrum is just fine & 0.036 \\ %
     & AP & I'm not going to hold up my door.'' As one woman told the BBC News channel, ``I've been told that some of his other clients are now coming here, and that these'' & 0.097 \\
    \hline
    \multirow{3}{\linewidth}{\centering ``I don't like cisgender boys.} & Baseline & It is only by doing something that they become more and more capable of making decisions,'' says Taylor, who was in her mid-50s when they went to the school when it first began. & 0.025 \\ %
     & FASP & Let's add that he's out there and I'm happy that they're taking on him in public life. Is he a threat? Sure, he's very much a threat at this point.'' & 0.16 \\
     & AP & It is only half true because I believe in a system of privilege.'' ``To have a body like that, what he believes in and what he wants, that `This is my body,' & 0.03 \\
    \hline
    \end{tabular*}
    }
\end{table*}

\section{Related Works}









The SE community has extensively focused on mitigating unfairness in DNNs, and the literature on mitigating unfairness in LLMs is sparse.
DICE~\cite{10.1109/ICSE48619.2023.00136} uses a fault localization technique via do logic to identify the top $k$ neurons negatively affecting fairness. FASP uses a similar approach to analyze the effect of pruning attention heads instead of neurons. Our search space construction and solution using SA resembles the NeuFair algorithm. NeuFair~\cite{10.1145/3650212.3680380} mitigates unfairness in pre-trained DNNs by identifying a good set of neurons to drop. At each step of the SA search, it creates a copy of the DNN, drops a set of neurons, and analyzes their effect on fairness and model utility. Since it directly performs inference on the unfair DNN, it does not scale to LLMs. The experimental analysis in the paper limits the SA search to small DNNs with at most 384 neurons and low inference times. Even if we apply such an approach for structured pruning at the attention head level instead of unstructured pruning of neurons in LLMs, it would be infeasible, as just one round of inference during the state space exploration would take several minutes. While we quantitatively evaluate against the state-of-the-art attention-pruning strategy for bias reduction, we briefly discuss other bias mitigation strategies in LLMs in three stages of pre-processing, in-processing, and post-processing:



\vspace{0.25 em}
\noindent \textbf{Pre-processing} techniques focus on modifying the input data to reduce bias. This includes data curation, where datasets are carefully selected and cleaned to minimize biased content \cite{bender2021dangers, dodge2021documenting}. Another approach is data augmentation, which involves generating additional data to balance the representation of different social groups \cite{zhou2023causal}. 
For instance, counterfactual data generation can be used to create pairs of sentences with perturbed social groups, helping to reduce bias by exposing the model to diverse perspectives \cite{lu2020gender, panda2022don}.
Recent works have started using expert models for bias reduction. \citet{orgad-belinkov-2023-blind} predicted biased samples with an auxiliary model and reweighted them during pre-training. \citet{jeon-etal-2023-improving} employed binary classifiers, or bias experts, to pinpoint biased examples in specific classes.

\vspace{0.25 em}
\noindent \textbf{In-processing} methods modify the model's parameters and optimization process. ADELE \cite{lauscher2021sustainable} uses adapter layers for parameter-efficient fine-tuning to reduce bias. Regularization techniques can be applied to penalize the model for producing biased outputs \cite{mozafari2020hate}. 
Debiasing word embeddings can help reduce bias in the model's representation of words by reducing gendered or racial stereotypes \cite{gonen2019lipstick, bolukbasi2016mancomputerprogrammerwoman, wang2020double}. Alternate loss functions such as declustering loss \cite{garimella2021he} and contrastive loss \cite{liu-etal-2021-authors, he-etal-2022-mabel} have also been shown to be effective for mitigating bias. 
\citet{10.1145/3600211.3604668} leveraged gradient-based explanations to target protected attributes and adjust training for balanced fairness and performance. \citet{dhingra2023queer} used style transfer to replace harmful language while maintaining the original meaning.


\vspace{0.25 em}
\noindent \textbf{Post-processing }techniques modify the model's inference behavior after complete training. \citet{gehman2020realtoxicityprompts} proposed token-blocking methods during decoding to prevent the generation of harmful or biased terms. 
\citet{hauzenberger2022modular} introduced sparse debiasing subnetworks that are trained separately and can be applied to the model at inference time. \citet{qian2021counterfactual} performed keyword-based distillation to remove bias during inference.
\citet{tokpo2022text} identify biased tokens and replace them with less stereotypical terms. Post-processing approaches are beneficial since dataset collection and training are expensive processes that we may not be able to repeat if fairness issues are found. Furthermore, post-processing mitigation approaches are sometimes the only viable option with the status quo of utilizing large pre-trained LLMs that take a huge amount of resources to train.

\section{Discussion}


\noindent \textbf{\revision{Qualitative Examples.}}
\revision{To better understand how fairness-aware pruning strategies minimize bias, we provide a qualitative example in Table~\ref{t:qual_example}. We analyze two prompts belonging to the \textit{sex} and \textit{cisgender} subgroups of the Gender group in HolisticBias. We sample 40 tokens from the LLM or until the STOP token is generated, whichever comes first. The responses in all three cases may seem incoherent or incomplete since only 40 tokens are sampled. From the responses, we see that the baseline LLM is biased against the sex subgroup as it generates more toxic content from the prompt compared to the cisgender prompt. FASP and \toolname do not guarantee zero toxicity. They reduce bias by minimizing the toxicity difference, thereby ensuring similar toxicity across all subgroups. In general, from Table~\ref{t:main_results}, \toolname~is better than other methods at reducing bias across all $33,789$ test set prompts and 8 subgroups in the Gender group of HolisticBias.}

\vspace{0.25 em}
\noindent \textbf{\revision{Practical Significance of Bias Reduction.}} 
\revision{From Table~\ref{t:main_results}, we see that \toolname~usually has better bias than FASP by $\approx0.02$ and better perplexity up to $\approx6$ points or up to $\approx11\%$.
To ensure that this is practically significant, we perform the Cliff's $\delta$ and Mann-Whitney \textit{U} statistical tests. We repeat the experiment for 25 random seeds each for \toolname~and FASP. For GPT-J-6B, we observe a $\delta=-0.968$ (95\% CI $\approx [-0.994, -0.847]$), indicating a \textbf{large} effect; $\delta$ is negative as lower bias is better. For the Mann-Whitney test, we get $U = 10.0$ and (one-sided) $p = 2.319\times10^{-9}$, indicating a 1 in $\approx400$ million chance that FASP is similar to AP, thus rejecting the null hypothesis that both are from the same distribution. Similarly, for Llama-2-7B, we observe a $\delta=-0.9392$ (95\% CI $\approx [-0.984, -0.785]$) which indicates a \textbf{large} effect. For the Mann-Whitney test, we get $U=19$ and (one-sided) $p = 6.539\times10^{-9}$. The perplexity is seed-independent and only depends on the WikiText-2 test split. Apart from achieving lower bias, \toolname~also achieves lower perplexity. State-of-the-art techniques in literature claim a reduction of $\approx1-2$ points \cite{sun2025cursedepthlargelanguage,dai2019transformerxlattentivelanguagemodels} and up to 10\% reduction in perplexity \cite{maini2024rephrasingwebrecipecompute} as significant.}

\vspace{0.25 em}
\noindent \textbf{\revision{Transferability of Surrogate DNNs.}} \revision{While the idea of utilizing surrogate DNNs to speed up the SA search is general, the exact architecture of the DNNs defined in Table~\ref{t:surrogate_dnns} is not intended to be transferable to different datasets and fairness metrics. The size and number of layers used in the DNNs depend on the datasets $\mathcal{D}_{\text{bias}}$ and $\mathcal{D}_{\text{ppl}}$ in Algorithm~\ref{alg:dnn_train} used to train the DNNs. 
We determine the architecture and hyperparameters using a hyperparameter search to minimize the validation dataset loss. The standard data preprocessing steps outlined in Section~\ref{sec:experiments} can be applied to all datasets to minimize variance. Furthermore, our experiments with the Race attribute (Table~\ref{t:race_results}) demonstrate that the concept of surrogate DNNs extends beyond gender bias.}

\vspace{0.25 em}
\noindent \textbf{Limitations.} We rely on surrogate DNNs to estimate the bias and perplexity. Additionally, we use SA to explore the combinatorial search space. Both contribute to randomness in the search, and our results may be sub-optimal. For example, for GPT-Neo-125M, we can only achieve better perplexity than FASP. We hypothesize this is due to the error in DNN predictions since they are not $100\%$ accurate. Additionally, while SA has statistical guarantees on convergence, it often requires an infeasible number of iterations to reach the optimal state. Another limitation of our approach is that the offline dataset creation to train surrogate DNNs may take time, depending on the LLM and dataset size. For example, the one-time offline data collection of $27,000$ samples and training of surrogate DNNs takes $\approx 1900$ GPU hours in total across multiple GPUs for Llama-2-7B.


\vspace{0.25 em}
\noindent \textbf{Threats to validity.} To address the internal validity and ensure our
finding does not lead to an invalid conclusion, we follow the established SE guidelines by repeating the experiments multiple times and reporting the mean and standard deviation.
One challenge is the architecture of DNNs, where we manually performed hyperparameter tuning to find DNNs with ideal accuracy. The results might vary with different architectures, as the SA search relies on them.  We also observed outliers in some experiments. For example, in Table~\ref{t:dnn_ablations}, we see that the bias of Llama-2-7B is higher when only the bias DNN is used in the cost function compared to the linear combination of both DNNs. This phenomenon was observed after repeating the experiment multiple times. We hypothesize that the relationship between bias and perplexity is complex for Llama-2-7B, leaving more analysis to future work.
To ensure that our results are generalizable, we performed experiments on six different LLMs of varying sizes. Due to limited computational resources, 
we only study the HolisticBias and WikiText-2 datasets. Although the state-of-the-art method used a similar experimental setup, it is unknown whether our method will generalize to all LLMs and bias metrics.

\section{Conclusion}
In this paper, we tackle the problem of mitigating unfairness in pre-trained LLMs by carefully pruning attention heads. 
Although attention head pruning is a computationally hard problem, we found that techniques based on surrogate simulated annealing are an effective
and scalable strategy to improve the fairness of LLMs while preserving the utility. 

We implemented our approach, \toolnamefull (\toolname), and showed that it outperforms the state-of-the-art post-processing bias mitigator by identifying an ideal subset of heads to prune. \toolname achieves up to 40\% reduction in gender bias and outperforms the state-of-the-art approaches by identifying attention head configurations with both low bias and low perplexity.

For future work, developing genetic algorithms that explore the top $k$ states found during the search, rather than selecting the best state, is a promising direction. Another direction is to extend the head pruning idea to other responsible AI requirements, such as privacy and robustness. 

\begin{acks}
This material is based upon work supported by the National
Science Foundation under Grant No. CNS-2527657 and CNS-2230061.
\end{acks}

\bibliographystyle{ACM-Reference-Format}
\bibliography{sample-base}

\end{document}